\documentclass[11pt]{article}
\usepackage{coling2020}
\usepackage{times}
\usepackage{url}
\usepackage{latexsym}
\usepackage{dirtytalk}
\usepackage{graphicx}	
\usepackage{subfig}
\usepackage{float}
\usepackage{array}
\usepackage{multirow}
\usepackage{xcolor}
\usepackage[toc,page]{appendix}
\usepackage{csvsimple}

\newcolumntype{?}{!{\vrule width 1.5pt}}
\newcolumntype{*}{!{\vrule width 2pt}}

\colingfinalcopy % Uncomment this line for the final submission

\title{Contextualized Word Embeddings Encode Aspects of Human-Like Word Sense Knowledge}

\author{Sathvik Nair, Mahesh Srinivasan\\
  Cognitive Science Program,\\ Department of Psychology,\\
  University of California, Berkeley\\
  Berkeley, CA, United States\\
  {\tt sathviknair,srinivasan@berkeley.edu} \\\And
  Stephan Meylan \\
  Department of Brain and Cognitive Sciences,\\
  Massachusetts Institute of Technology \\
  Cambridge, MA, United States\\
  {\tt smeylan@mit.edu}}

\date{}

\begin{document}
\maketitle
\begin{abstract}
Understanding context-dependent variation in word meanings is a key aspect of human language comprehension supported by the lexicon.
Lexicographic resources (e.g., WordNet) capture only some of this context-dependent variation; for example, they often do not encode how closely senses, or discretized word meanings, are related to one another.
Our work investigates whether recent advances in NLP, specifically contextualized word embeddings, capture human-like distinctions between English word senses, such as polysemy and homonymy.
We collect data from a behavioral, web-based experiment, in which participants provide judgments of the relatedness of multiple WordNet senses of a word in a two-dimensional spatial arrangement task.
We find that participants' judgments of the relatedness between senses are correlated with distances between senses in the BERT embedding space.
Homonymous senses (e.g., bat as mammal vs. bat as sports equipment) are reliably more distant from one another in the embedding space than polysemous ones (e.g., chicken as animal vs. chicken as meat).
Our findings point towards the potential utility of continuous-space representations of sense meanings.
\end{abstract}
\section{Introduction}
\blfootnote{
    % % final paper: en-us version 
    %
    \hspace{-0.65cm}  % space normally used by the marker
    This work is licensed under a Creative Commons 
    Attribution 4.0 International License.
    License details:
    \url{http://creativecommons.org/licenses/by/4.0/}.
}
A key challenge in natural language understanding is grasping the range of meanings that a word can take as a function of linguistic and non-linguistic context.
Successful linguistic comprehension involves constantly resolving lexical ambiguity of this nature \cite{frazier1990taking,klepousniotou2002processing}.
This re-use of word forms by speakers --- relying on listeners to choose the appropriate meaning depending on context --- confers language with higher communicative efficiency than it would otherwise have, and may pose a solution to the problem of limited memory \cite{piantadosi2012communicative}.
The re-use of word forms also allows speakers to extend the lexicon to new communicative situations, for example to refer to new objects, entities or processes using existing words in the language \cite{ramiro2018algorithms,xu2020conceptual,srinivasan2019children}.
How these context-specific meanings are understood and represented pose critical open questions regarding the lexicon.

Following the conventions of lexicographic resources like natural language dictionaries, variation in word meaning is often treated in a categorical fashion: lexical types contain clusters of related meanings, or \textit{word senses}. 
Under this treatment, word tokens (instances of a word type that are used in context) can be categorized into these word senses.
Different sub-types of lexical ambiguity are thus reflected in different relations between word senses:
pairs of word senses are \textit{polysemous} if they are semantically related \cite{pustejovsky1998generative}, or \textit{homonymous} if this is not the case \cite{apresjan1974regular,tuggy1993ambiguity,lyons1995linguistic}.
For example, using \say{bottle} to refer to a container as well as to the liquid it contains is an example of polysemy.
By contrast, using \say{bank} to refer to a riverside or a financial institution --- two semantically unrelated meanings --- constitutes homonymy.
Polysemous relations are often \textit{regular} \cite{apresjan1974regular}, such that many word types exhibit the same alternation (e.g., the container-for-contents relation is also exemplified by \say{box} and \say{glass}); such patterns can also be generalized to new words \cite{srinivasan2019children}.

Because of variation in the perceived relationship between senses across language users, theories in psycholinguistics assert that humans treat polysemy and homonymy as falling onto a continuous gradient \cite{tuggy1993ambiguity,crossley2010development}.
This is corroborated by psycholinguistic experiments which show that both adults and children represent and process polysemous and homonymous senses differently \cite{frazier1990taking,rodd2004modelling,klepousniotou2008making,rabagliati2013truth,macgregor2015sustained}.
However, these theories of word sense representation lie in stark contrast with many lexicographical approaches, most notably WordNet \cite{miller1990introduction}.
Although WordNet contains a vast store of word senses---and has historically been considered the gold standard ontology of word senses \cite{jurafsky2014speech}---it does not encode relations among those senses, and thus does not distinguish between polysemy and homonymy. 

Models based on distributional semantics can help bridge the gap between psychological theories of word senses and existing lexicographical resources. 
Contextualized word embeddings (CWEs), especially BERT \cite{devlin:2018}, which is derived from a Transformer-based architecture \cite{vaswani2017attention}, have the potential to capture fine-grained distinctions in relatedness between word senses, because they offer a continuous measure of relatedness between individual uses of words in context \cite{lake2020word}.
While fine-tuned BERT models are known to perform exceptionally well in word sense disambiguation (WSD) tasks  \cite{wiedemann2019does,loureiro-jorge-2019-language,blevins2020wsd}, there are still open questions of whether BERT representations capture fine-grained distinctions in sense relatedness.%critical open qs...in the relatedness of different senses.

In the present work, we evaluate whether contextualized embeddings from BERT capture relationships between word senses similar to English speakers. % as described by people. 
We collect human judgments of the relatedness among different senses of 32 English words through a web-based two-dimensional spatial arrangement task \cite{goldstone1994influences}.
We then compare the experimental data to BERT vectors for the same set of words.
This is done through extracting and analyzing BERT embeddings for word tokens in the Semcor corpus \cite{miller1993semantic}, which has been annotated with word sense identifiers from WordNet.
Assessing whether relatedness in BERT representations corresponds with human judgments of sense relatedness is an important test of whether CWEs can be used to develop more realistic computational models of word meanings.

\section{Background}
% Figure out where to place: People often use linguistic context to effectively deduce the sense of a word \cite{Mcdonald01testingthe}; computational models have been developed to imitate this ability.
%However, they do not represent homonymy and polysemy accurately \cite{arora2018linear}, as all the senses for a word are represented as a single vector without further architectural elaboration (e.g., Sense2Vec, \cite{DBLP:journals/corr/TraskML15}).

The distributional hypothesis proposes that a word's meaning can be represented by the lexical context in which it occurs \cite{harris1954distributional}.
This insight underlies the success of computational models that represent words as continuous-valued vectors representing their surrounding lexical context.
Latent semantic analysis \cite{dumais1988using}, and neural models like Word2Vec\cite{mikolovw2v} encode information about all of a word's senses in a single vector. %we say "recent neural models" in the next sentence too
% \footnote{Certain methods, such as Sense2Vec, \cite{DBLP:journals/corr/TraskML15}), elaborate on the architecture of models like Word2Vec, but do not represent word senses in a manner that can be categorized like WordNet does.}
By contrast, newer models include contextualized embeddings: vector space representations of specific word uses (tokens), reflecting the context of their use.
%By contrast, recent neural network models learn representations for every word token in a dataset, or contextualized word representations (CWEs).
%These models output different vectors for the same word type when used in different contexts. 
These richer word representations from contextualized embeddings have allowed researchers to ask which aspects of linguistic knowledge are incorporated in their vector spaces (See \newcite{rogers2020primer} for a review; see also \newcite{ethayarajh-2019-contextual}).
For example, one of the most notable discoveries from this line of work suggests that a sentence's dependency parse tree can be reconstructed from BERT embeddings \cite{hewitt2019structural,reif2019visualizing,jawahar-etal-2019-bert}. 
However, despite progress investigating how BERT encodes syntactic information, relatively little work has explored how it encodes semantic information such as the relations between word senses.

While older models like Word2Vec may encode some aspects of word sense \cite{arora2018linear}, we focus here on Transformer architectures \cite{vaswani2017attention}, which may be able to find sense-distinguishing context words at larger distances, and thus perform better for WSD. 
BERT-based approaches have led to state of the art performance on word sense disambiguation tasks \cite{wiedemann2019does,loureiro-jorge-2019-language,blevins2020wsd}, but approaches analyzing the model's representation of word senses remain exploratory \cite{reif2019visualizing,lake2020word}.
%"words of the same sense would be closer together and words of different senses would be further apart," is the same thing true of the relationship between WN senses?
Evidence from \newcite{reif2019visualizing} shows that BERT embeddings for the same word in a large text corpus are largely clustered based on their meanings, but the authors do not investigate the model's ability to encode canonically homonymous or polysemous relationships.
\newcite{mickus2019you} note that BERT embeddings for the same word type vary as a function of the position in the sentences they occur in.
\newcite{lake2020word} assess BERT's capacity for homonym resolution by comparing relatedness of words that are similar to a homonym in highly constraining contexts.
\newcite{ettinger2020bert} points out the need for more psycholinguistic diagnostics of neural language models like BERT, and compares BERT to human data from tasks such as commonsense inference. 
Our work falls under this framework, focusing on comparing human judgements of word sense relatedness to BERT representations.
\section{Methods}
We conducted a metalinguistic experiment (vs. a processing task testing implicit knowledge) where participants used a web interface to assess the relatedness of WordNet senses in a two-dimensional spatial arrangement task.
We then obtained BERT embeddings for word types in the Semcor corpus \cite{miller1993semantic} and compared them to the experimental data, looking at both cosine distance in the embedding space as well as the accuracy of a sense classifier using BERT's contextualized word embeddings as input.
All stimuli, code and visualizations are available at \url{https://osf.io/fm78w}.
\vspace{-3mm}
\subsection{Data}\label{data_stimuli}
We select a sample of word types for analysis from Semcor \cite{miller1993semantic}, which has WordNet sense annotations for 235,000 tokens from the Brown corpus \cite{francis1979brown}. 
We use the corpus reader from the Python Natural Language Toolkit (NLTK) \cite{bird-loper-2004-nltk} to access the corpus.
The fact that syntax is represented within BERT embeddings \cite{hewitt2019structural} implies that BERT embeddings can easily capture distinctions in part of speech, so we focus on lexical ambiguity within part-of speech (specifically nouns and verbs) as a more challenging test case.\\

For the behavioral experiment, we selected word types across a range of sense entropy values (i.e., uniform to highly peaked sense distributions).
We created a multinomial distribution over the senses of each lemma, i.e., (word type, part of speech) pair, and computed its entropy as follows:
$\displaystyle -\sum_{s \in L}\frac{c_s}{c_L} \log (\frac{c_s}{c_L})$, where $L$ is a lemma, $s$ is a sense and $c$ corresponds to a frequency, or count in the corpus.
We removed stopwords and lemmas with zero entropy (79.3\% of all lemmas).
This yielded 444 word types. %, and generated embeddings associated with the tokens for  with BERT (further discussed at the beginning of Section \ref{modeling}).\\ 

For experimental stimuli, we chose 32 word types.
20 word types came from Semcor, including 11 low/medium entropy word types and 9 high entropy word types.
We defined high entropy to be greater than 1.5 when rounded to the nearest tenth, and medium/low entropy to be less than this value, based on the distribution of sense entropy in the available Semcor data (in Figure \ref{fig:f1_entropy}).
To account for variability in participants' placement of tokens in the spatial arrangement task, we needed to normalize measurements in the interface on a per-trial basis, so we selected words with three senses or more.
%In order to measure the effect of a lemma's sense distribution's complexity, 
We chose word types with varying entropy values to determine if this quantity had an effect on the correlation between BERT vectors and human judgements; relatedness between embeddings for words with more unpredictable sense distributions may be less consistent with results from the experiment.
To avoid overwhelming participants, we selected words with fewer than eight senses.
We also chose six additional words of theoretical importance that exemplify patterns of regular polysemy that have been observed across languages from \newcite{SrinivasanRabagliati15}, and manually chose six words with three senses, one of which was less semantically related to the other two.
Because we expected these words to exhibit the greatest differences in pairwise similarity, we elicited judgments from all participants on these items.
More details about how these stimuli were presented and exclusion criteria are in Appendix \ref{trialtypes}.
%For some of these word types, we included senses with four or more instances rather than 10, because we needed to provide participants with a third sense as mentioned above.
%Once we found the 32 word types, we collected definitions of the WordNet senses covered by Semcor, as well as example sentences.
%We describe the order in which these words were presented to participants in Section \ref{trialtypes}.

\subsubsection{Participants}
105 undergraduate students from a major research university participated in the experiment and were compensated with class credit. 
Upon providing informed consent, participants reported their experience with English and other languages to ensure that data were collected from proficient English speakers (defined as at least 50\% of daily language use); one participant was excluded using this criterion.
%Additional exclusion criteria are described in Appendix \ref{trialtypes}. 

\subsection{Experiment}
From the set of 32 word types, each participant received 14 word types as stimuli.
We used a two-dimensional spatial-arrangement task \cite{goldstone1994influences}, because it allowed us to efficiently capture psychological judgments, and because tasks of this nature effectively capture relatedness in a high-dimensional semantic space \cite{richie2020spatial}.
Participants were told to place less related sense tokens further apart from one another on the canvas, and more closely related tokens near each other.
The experimental interface is shown in Appendix A.
To place the sense tokens, participants were given definitions (from WordNet) and example sentences (from Semcor; in some cases shortened for brevity). %not manually shortened all the time
The task was untimed, and participants were encouraged to introspect for as long as necessary about the meanings of the presented senses before making their placements.
Partcipants were encouraged to adjust their placements to reflect all senses.

\subsection{Relatedness Matrices}
For each word, we collected spatial (x,y) positions of each of the word's senses from each participant.
To derive an estimate of the relatedness of senses for each word type, we normalized the distances such that they were rescaled according to the largest reported pairwise distance. 
This controlled for variation in the absolute amount of space in the canvas that different participants used when making their placements. After excluding participants with unreliable responses (Appendix \ref{trialtypes}), we then averaged relatedness matrices for each word type across the remaining participants.
This yielded a single \say{aggregate relatedness matrix} for each word type which could then be directly compared with model results for a word type.
To account for the larger number of participants reporting data for shared stimuli, we selected data from a random set of participants whose size corresponded to the average amount of test items ($n$ = 29).
Data from a minimum of 21 participants and a maximum of 37 participants were used to construct the aggregate relatedness matrix for each word type.

\subsection{Modeling}\label{modeling}
We retrieved BERT embeddings for labeled tokens from the Semcor corpus \cite{miller1993semantic} and ran classifiers to distinguish between different senses of individual words based on BERT vectors.
We compared these data to results from the experiment based on two metrics: distance in the embedding space and classification accuracy, assessing their similarity through correlation \cite{Hirst06evaluatingwordnet-based}.

Given a word type and a part of speech, we derived word embeddings from BERT as follows.
For each sense of the (word type, part of speech) pair, we retrieved sentences from Semcor, and tokenized them using rules specified by the BERT authors \cite{devlin:2018}.
%such as padding sentences with start and stop tokens.
We loaded a pre-trained BERT model \texttt{BERT-base} from \newcite{Wolf2019HuggingFacesTS} and ran the forward pass on each sentence, extracting the activations corresponding to the type, and storing the summed activations of the final four layers, which has produced strong results in word sense disambiguation \cite{loureiro-jorge-2019-language}. 

\subsection{Word Sense Classifier}\label{wsd}
As a matter of due diligence, we first evaluated the effectiveness of BERT representations for word sense disambiguation (classification) in the Semcor corpus. 
While WSD-focused work has used the Semcor dataset for training and tested on other datasets \cite{raganato2017word}, we confirmed that BERT demonstrates similar levels of performance on Semcor. %delete wanted to
To this end, we assessed performance of multiclass logistic regression models to predict word senses for 401 word types from Semcor.
Senses were omitted from this classification whenever there were fewer than 10 tokens.
Each logistic regression model was trained to predict WordNet sense labels for each instance of a sense based on its contextual word embedding from BERT. %uses data from the word's BERT embeddings as \todo{features} is redundant
We conducted 5-fold cross validation and applied L1 regularization to prevent overfitting.
To confirm consistency with prior work on WSD \cite{loureiro-jorge-2019-language,reif2019visualizing,blevins2020wsd}, we report the average F1 score (weighted for each sense) across the runs for each word type.
Comparable F1 scores relative to this prior work would suggest that it is appropriate to further investigate the geometry implicit in these contextual word embeddings.

\subsubsection{Extracting Cosine Distances Between Word Sense Centroids}\label{cosine}
After qualitatively reproducing previous word sense disambiguation results  using BERT embeddings, we then tested whether the relationship between BERT embeddings of senses parallels human judgments of relatedness between senses.
Because BERT generates one vector for each use of a word token, we computed the centroids of the BERT embeddings corresponding to each sense, and then compared different senses by using the cosine distance of these centroids to one another.
In a broad range of models, cosine distances between vectors corresponding to word types can encode the degree of semantic relatedness \cite{dumais1988using,mikolovw2v,joulin2016fasttext}, so we can use this same metric with BERT embeddings to compare word senses to one another.
Following the same procedure as the relatedness matrices, we define relatedness as (1 - cosine distance), such that the largest cosine distance (least related pair) takes the value of 0 and the smallest cosine distance (most closely related pair) takes the value of 1 for each word type.
We stored these relatedness measures between sense centroids in a matrix analogous to the aggregate relatedness matrices from the behavioral experiment.
To evaluate their fit to human data, we computed the Spearman rank correlation coefficient between the upper triangular entries in these two distance matrices.
A nonparametric measure of correlation is appropriate given that the relationship is not necessarily linear.

\section{Results}
First, we verified that BERT can discriminate between senses in Semcor comparably to other test datasets.
We then compared relatedness estimates derived from distances in the BERT embedding space to the aggregate relatedness measures from the experiment.
We then conduct two exploratory analyses, the first comparing the relatedness of homonymous and polysemous sense pairs, and the second examining the utility of pairwise sense confusion as a measure of relatedness.
% SM: I think we are going to want to reword this if we can test pairwise homonymous vs. pairwise polysemous relations 
\subsection{Classification Results for All Semcor Data}\label{wsd_results}
\begin{figure}
    \centering
    \includegraphics[scale = 0.6]{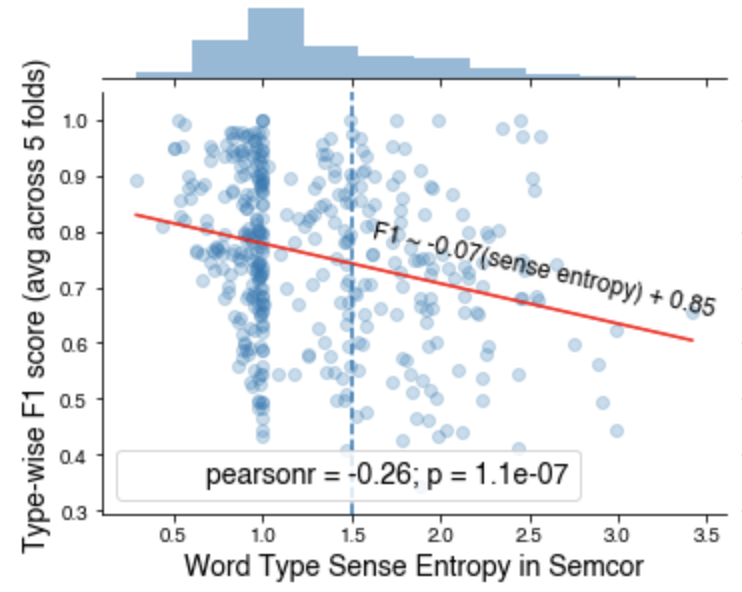}
    \caption{Classifier performance for word types with respect to their sense entropy in Semcor. F1 score for each type is averaged over cross validation folds and weighted by sense frequency. Vertical line represents boundary between low/medium and high entropy word types.} %\todo{this under/over line in figure: We also fit a linear model ($\beta_{entropy} = -0.129, R^2 = 0.122, intercept = 0.9$). Pearson's $r = -0.350, p < 0.001$}}
    \label{fig:f1_entropy}
\end{figure}

We compared the distribution of F1 scores of the classifiers on the 401 words from Semcor (BERT failed to process 43 of the words) with two baselines: random choice and selecting the most frequent sense (Figure \ref{fig:logistic_semcor}).
\begin{table}
    \centering
    %\vspace{-0.3cm}
    \begin{tabular}{|c|c|c|}
    \hline
        Method & All Types in Semcor & Types in Behavioral Expt.\\\hline
        Number of Available Types & 401 & 32\\
        Random Sense & 0.480 & 0.423\\
        Majority Sense & 0.441 & 0.403\\
        \textbf{Logistic Regression on BERT Embeddings} & \textbf{0.757} & \textbf{0.797}\\
        \hline
    \end{tabular}
    \caption{Average F1 scores (across types and train-test splits) for BERT-based word sense disambiguation in the Semcor corpus.
    \vspace{-5mm}}
    \label{fig:logistic_semcor}
\end{table}
Classification performance on all 401 words, including 25 words used in the behavioral experiment, were similar to those reported by \newcite{blevins2020wsd}, \newcite{loureiro-jorge-2019-language}, and Reif et al. \newcite{reif2019visualizing} (0.739, 0.754, and 0.711, respectively), which used test sets from \newcite{raganato2017word}.
We find that type-wise F1 score is negatively correlated with sense entropy computed from Semcor ($\beta_{sense\_entropy} = -0.074, R^2 = 0.06$; Fig. \ref{fig:f1_entropy}).

\subsection{Comparing Human Relatedness with BERT Word Sense Geometry}\label{corr_hb}

We computed the Spearman's rank correlation coefficient between human judgments of sense relatedness (from the aggregate relatedness matrices)  and measures of word sense relatedness from the centroids of BERT contextualized word embeddings.
This yielded a positive correlation (Spearman's $r = 0.565, p < 0.001$), with a 95\% confidence interval of (0.459, 0.655).
%Because of the varying number of samples for certain stimuli as mentioned in Section \ref{cosine}, we plot a distribution of correlations after resampling 1000 times,
To confirm that this correlation was above that expected by random chance, we compared it to a random baseline established by 1000 draws of randomly generated sense placements for 29 participants (the average number of participants per item). 
This correlation was much lower (Spearman's $r = 0.062, p < 0.001$,  95\% CI: -0.065 - 0.208).
We report the correlation between relatedness measures from the embedding space and human judgements by part of speech and high vs. low/medium entropy in Table \ref{tab:cosine_conf_corrs}.
Correlations between human relatedness judgments and BERT relatedness were substantially higher among low to medium than high entropy words(Table \ref{tab:cosine_conf_corrs}).
We also find that correlations among human and model relatedness are higher among verbs than nouns.

\subsubsection{Polysemous vs. Homonymous Sense Pairs}

We conducted an exploratory analysis to see if human relatedness judgments differed between homonymous and polysemous relations, and whether the geometry of BERT embeddings captured this distinction between word senses.
In the absence of gold-standard datasets labeling polysemous and homonymous word sense relationships,\footnote{We could in principle use natural language dictionaries, some of which list polysemous and homonymous senses under separate entries. However, such definitions would not necessarily map to WordNet labeled data.} we tagged pairs of senses as polysemous or homonymous ourselves for the set of words in the behavioral experiment.
For human judgments, we took the set of average distances between polysemous and homonymous pairs of senses.
%and aggregate \todo{over homonymous and polysemous} pairs.
For BERT embeddings, we took the cosine distances between pairs of polysemous and homonymous sense centroids. 
%and aggregate by the same set of labels.

\begin{figure}
\centering
\includegraphics[scale = 0.38]{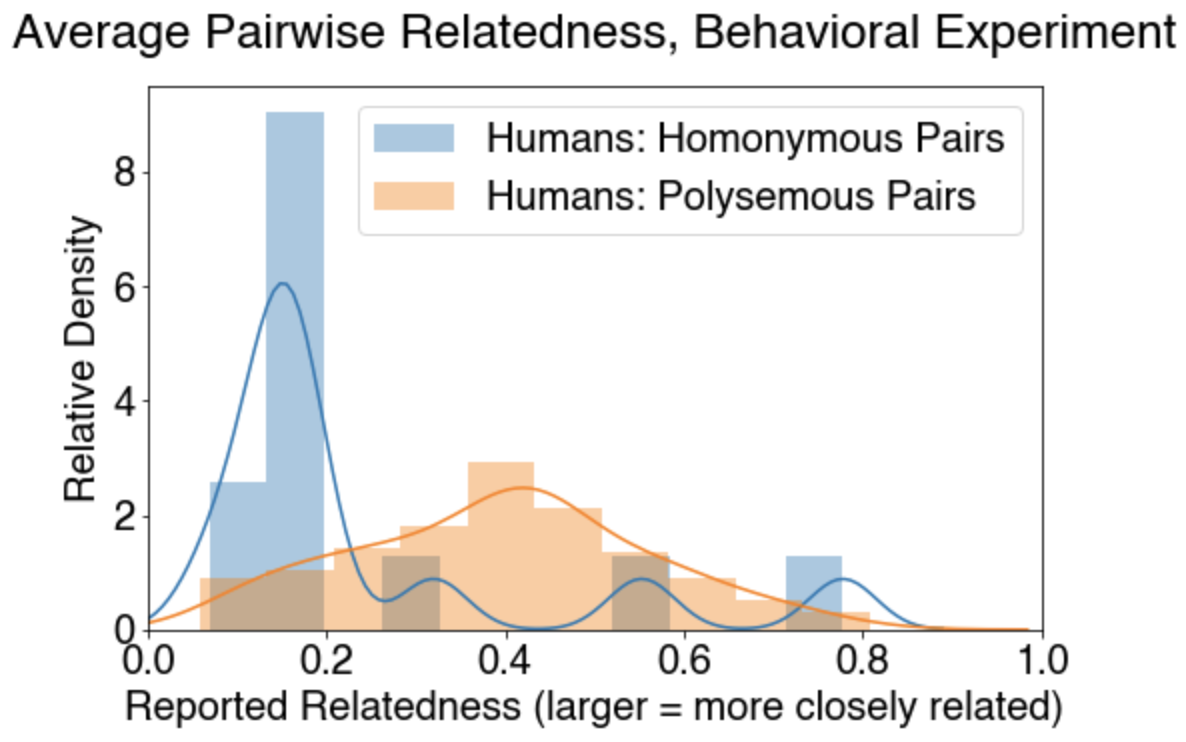}
\includegraphics[scale = 0.38]{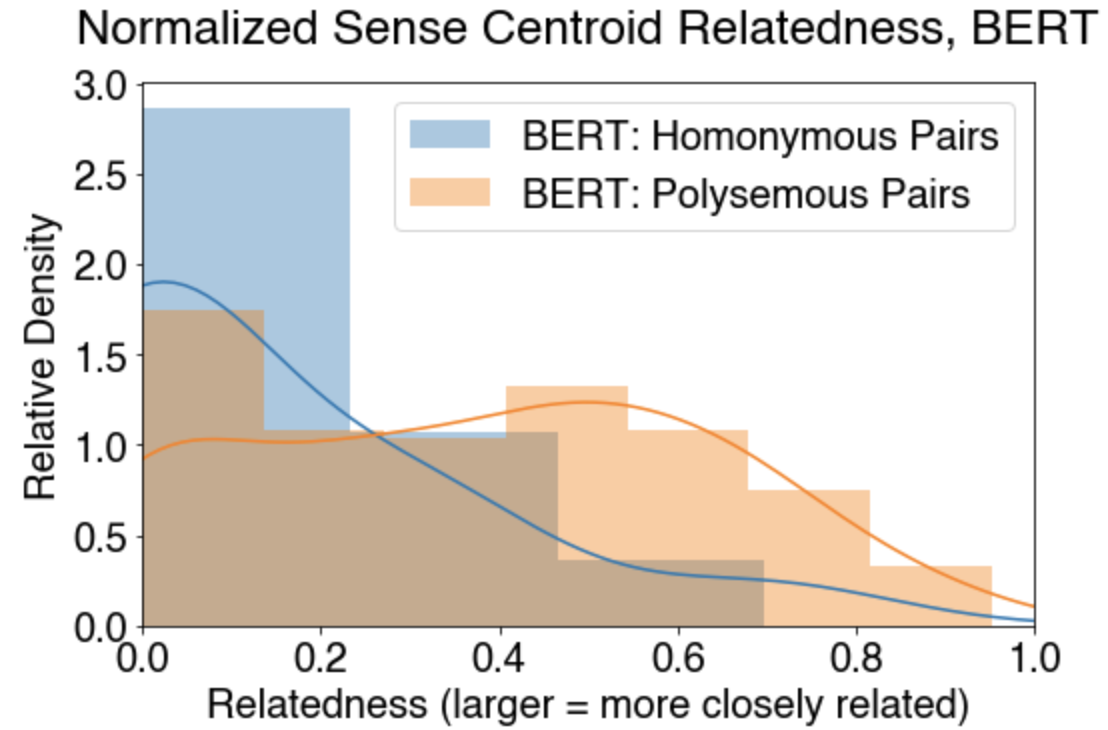}
\caption{Distribution of distances between pairs of senses from the human data (left) and from BERT (right), across word types.}
\vspace{-0.3cm}
\label{fig:pairwise}
\end{figure}

The density plots in Figure \ref{fig:pairwise} show that participants in the behavioral experiment judged the polysemous pairs to be more similar (i.e., less distant) than homonymous pairs (Mann-Whitney $U = 516, p < 0.001$).
BERT-based representations reproduced this basic pattern (Mann-Whitney $U = 493, p < 0.001$), though many relatedness estimates were more dispersed for both polysemous and homonymous pairs.

\begin{figure}[t!]
    \centering
    \includegraphics[width=\textwidth]{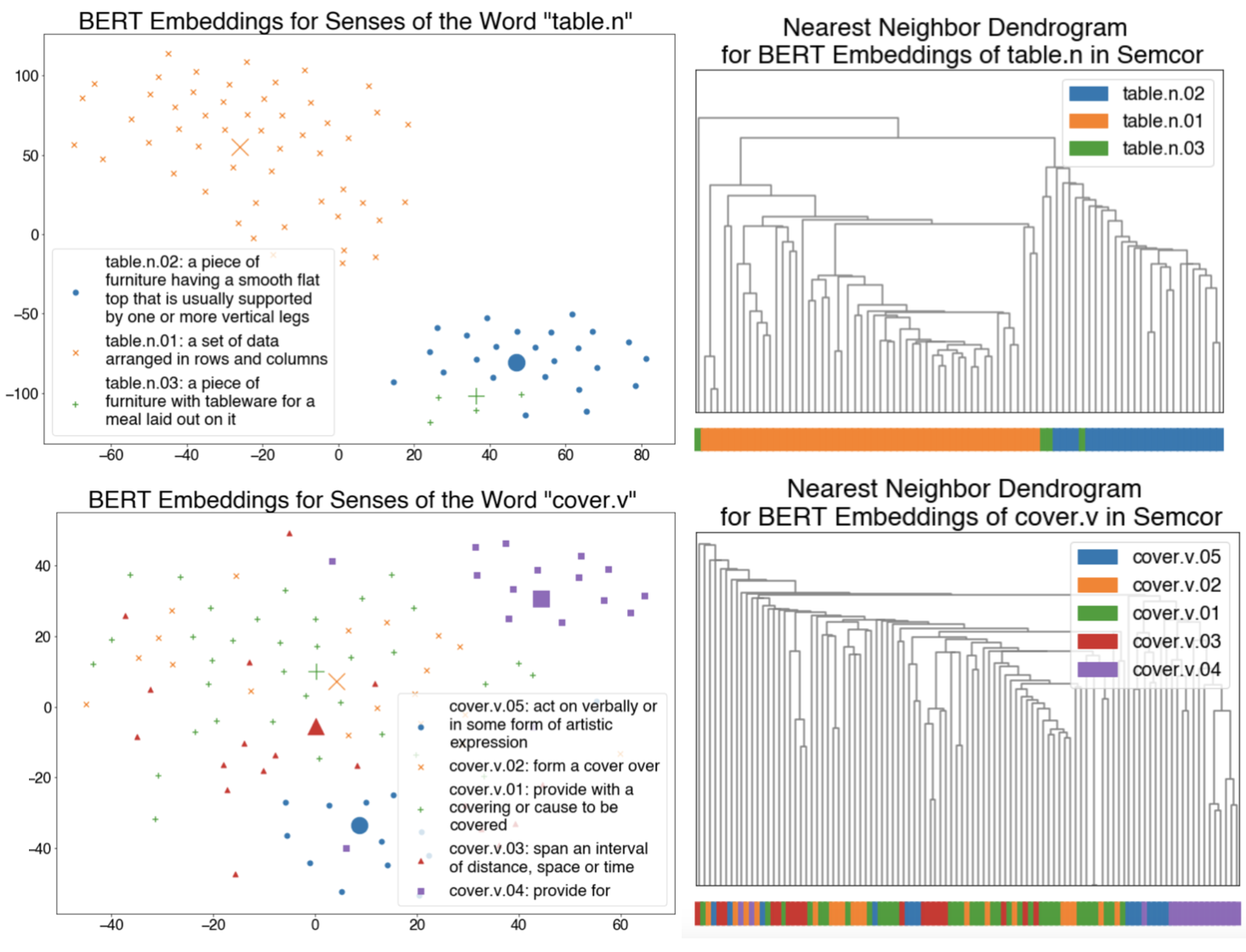}
    \caption{Visualizations of example BERT embeddings for stimuli \texttt{table.n} \texttt{cover.n}. Left column contains results of t-SNE dimensionality reduction with small points indicating word tokens and large points indicating sense centroids. Right column contains results of single linkage agglomerative clustering.}
    %\vspace{-1cm}
    \label{fig:bertviz} 
\end{figure}

To illustrate the differences between how BERT represents polysemous and homonymous relationships, we show t-SNE and dendrogram visualizations in Figure \ref{fig:bertviz}.
For \texttt{table.n} (Figure \ref{fig:bertviz}, top), we show that two senses related by polysemy (piece of furniture vs. a tablesetting) are judged to be closer together than homonymous pairs (either of the first two senses vs. ``a set of data arranged in rows and columns'').
For \texttt{cover.v}, all sense pairs were polysemous. 
Instances of the same sense were still closer to one another in the embedding space, but tokens corresponding to different senses were much less clearly distinguished except for \texttt{cover.v.04}, which may be well-separated from other senses because it refers to a metaphorical meaning of the lemma.
We find that the BERT centroids' cosine distances and the aggregated relatedness judgements are strongly correlated (Spearman's $r = 0.851, p < 0.001$), with a 95\% confidence interval of (0.638, 0.943).

\subsection{Predicting Human Relatedness Judgments with Pairwise Confusion}

One possibility is that the cosine distance computed over all dimensions of the BERT embedding space overlooks the possibility that some dimensions may be more or less useful for discriminating word senses depending on the word type.
One alternative is to use the pairwise confusion probabilities from multiclass logistic regression (described in Section \ref{wsd}) as a measurement of pairwise relatedness in the model.
This choice invokes substantive theoretical questions: in principle, sense relatedness and discriminability could be orthogonal.
For example, two senses could be judged as closely related by humans, but they could in principle be able to discriminate between instances of those senses without errors. %a population of human raters
In the absence of human sense discrimination performance data, we leave this question to future work.
%delete: to link with the relatedness judgments collected here
Nonetheless, we investigated the utility of pairwise sense confusion as an alternative predictor for human judgments of sense relatedness.

\begin{figure}

\subfloat[Subfigure 1 list of figures text][]{
\includegraphics[width=3.5cm,height=4cm]{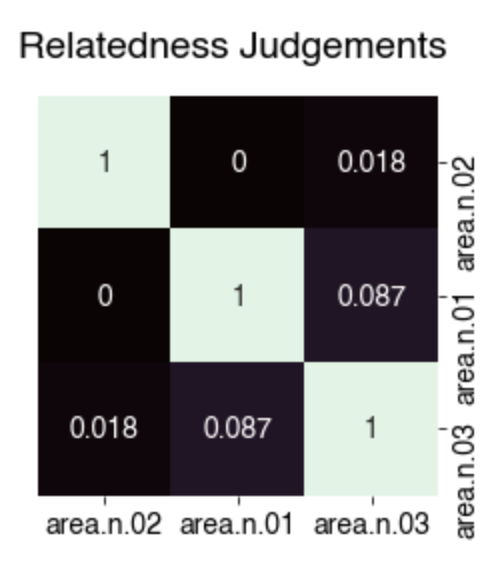}
\label{fig:subfig1}}
\subfloat[Subfigure 2 list of figures text][]{
\includegraphics[width=4.2cm,height=3.9cm]{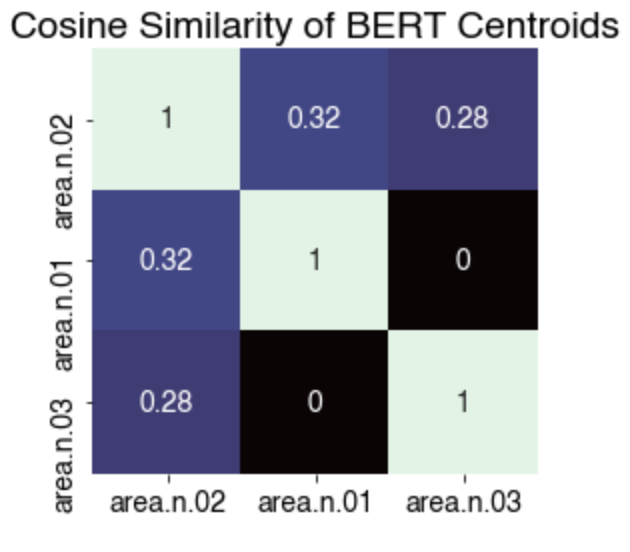}
\label{fig:subfig2}}
\subfloat[Subfigure 3 list of figures text][]{
\includegraphics[width=4.2cm,height=3.9cm]{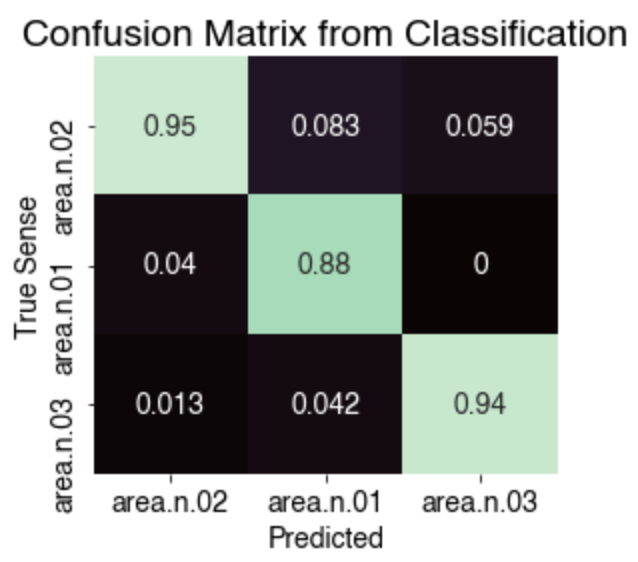}
\label{fig:subfig3}}
\subfloat[Subfigure 4 list of figures text][]{
\includegraphics[scale=0.5]{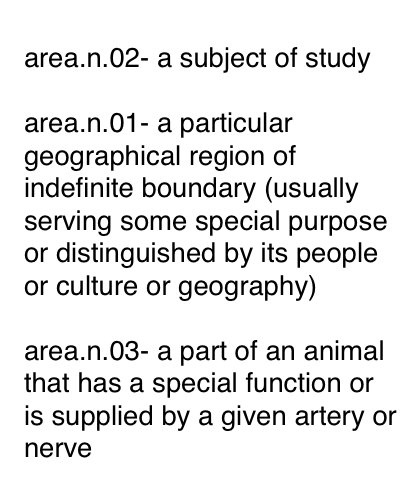}
\label{fig:subfig4}}

%\small
%\begin{tabular}[h]{|c|p{2cm}|}\hline
 %   area.n.01 & a particular geographical region of indefinite boundary \\\hline
 %   area.n.02 & a subject of study\\\hline
 %   area.n.03 & a part of an animal that has a special function or is supplied by a given artery or nerve\\\hline
 %   \end{tabular}

    \caption{Mean human relatedness judgements, cosine similarity matrix, and confusion matrix for senses of \texttt{area.n}.}
    \label{fig:area}
    \vspace{-0.25cm}
\end{figure}

To this end, we summed the confusion matrices when senses from the stimuli are classified during each iteration of cross-validation from Section \ref{wsd_results}.
We normalized each item in the matrix by the number of true labels, such that it represents the probability an item was predicted given its true class.
Across all word types with available data (20 out of 32 types, including 150 out of 189 sense pairs), we found a positive Spearman's rank correlation between entries in the confusion matrices and matrices of the corresponding relatedness judgements $(r = 0.649, p < 0.001)$, with a 95\% confidence interval of (0.592, 0.7).
For this measure, we considered all entries in both matrices, as the confusion matrices are not necessarily symmetric.
Because this analysis omits word senses with fewer than 10 tokens in Semcor, this evaluation reflects a smaller set of items than the one reported in Section \ref{corr_hb} for cosine-based relatedness.
When we computed the correlation between cosine-based relatedness and human aggregated relatedness matrices on this smaller set, the correlation remains lower (Spearman's $r = 0.518, p < 0.001$), with a 95\% confidence interval of (0.391, 0.627). 
This suggests that pairwise confusion in BERT may be a stronger predictor of human relatedness judgments.
We also report correlations stratified by part of speech and type sense entropy in Table \ref{tab:cosine_conf_corrs}.
\begin{table}
    \centering
    \begin{tabular}{|c*c|c?c|c|}\hline
        \multirow{2}{*}{Metric}& \multicolumn{2}{|c|}{Part of Speech} & \multicolumn{2}{*c|}{Word Sense Entropy} \\ \cline{2-5}
         & Nouns & Verbs & High & Low/Medium \\\hline
        Relatedness (1 - Normalized Cosine Distance) & 0.487 & 0.623 & 0.345 & 0.570\\\hline
        Pairwise Confusion  & 0.609  & 0.627 & 0.518 & 0.731\\\hline
    \end{tabular}
    \caption{Spearman rank correlations between relatedness judgements and confusion matrices in BERT embedding space, split over part of speech and entropy level}
    \label{tab:cosine_conf_corrs}
\end{table}
To illustrate the approach, we present aggregate relatedness matrices and confusion matrices for the word type \texttt{area.n} in Fig. \ref{fig:area}.  % and \texttt{indicate.v}  (Figures \ref{fig:area}, \ref{fig:indicate}).
In this case, the experimental data (Fig \ref{fig:area}, A) was more closely aligned with the confusion matrix (C)  the than the cosine similarity of centroids (B).

We also evaluated sense classification accuracy on the basis of the homonymous and polysemous sense pairs from the preceding exploratory analysis. 
%%SN- we only had enough instances of the homonymous sense pairs
Among homonymous sense pairs, the classifiers achieved an average F1 score of 0.992.
Among the polysemous sense pairs, the classifier achieved an average F1 score of 0.752.

\section{Discussion}

We investigated the ability of an artificial neural network model that represents word tokens with contextualized word embeddings to capture human-like relations among English word senses.
On a subset of word types from the Semcor corpus, we reproduced previous word sense disambiguation results.
We then showed that these same BERT embeddings capture a significant amount of information regarding the relationship between word senses, and are able to at least partially reproduce human relatedness judgments.
An exploratory analysis revealed that BERT-based measures of the relatedness between pairs of homonymous senses are much lower than for pairs of polysemous senses, matching human intuitions for the same set of senses.
Pairwise confusion from the sense classifiers provided a slightly better predictor of human judgments of sense relatedness compared to distance in the embedding space, although this may be unsurprising given that pairwise confusion reflects some additional degree of supervision.
Analyzing the error rates of classifiers trained on BERT embeddings of the stimuli, we also provided evidence that their results are more accurate for homonymous sense pairs compared to polysemous sense pairs.\\

Following in the approach of comparing BERT vectors to experimental data from human participants \cite{ettinger2020bert}, the present research addresses how BERT represents word senses, bridging the gap between human and computational models of lexical ambiguity resolution. 
Recent progress showing BERT's performance on word sense disambiguation \cite{wiedemann2019does,loureiro-jorge-2019-language,blevins2020wsd} indicates that BERT-based models perform better than past approaches, and our work specifies that the representations they use to accomplish this are relatively consistent with human intuitions.
Existing exploratory work aims to analyze how BERT represents word senses \cite{reif2019visualizing,lake2020word}, but we systematically evaluate these claims with both WordNet senses and human data.
Our work corroborates claims by \newcite{lake2020word} that BERT captures relations between homonymous senses.
Indeed, poor performance in discriminating polysemous pairs of words senses suggests that more work needs to be done on capturing polysemous relations; one possibility is to explore how BERT could be fine-tuned to capture relations exemplified by regular polysemy.
One limitation from these findings is that the dataset only covers 32 word types, which we hope to address in future work.

More generally, our findings suggest that a continuous measure of sense relatedness derived from neural language models could potentially be used to augment existing lexicographic resources like WordNet, in this case providing a measure of relatedness between senses.
We encourage other researchers developing sense ontologies, especially those found through word sense induction, to consider representing the relationships among word senses in addition to discovering new sense inventories.

\section{Conclusion}
Through comparing results from a behavioral experiment with data from contextualized word embeddings, we demonstrate that new Transformer-based neural network architectures may reflect human intuitions about the relationships between word senses, at least in English.
Exploratory analyses suggest that these measures of relatedness between sense pairs reflect a distinction between polysemous and homonymous relationships between word senses, but that these models are much more effective in discriminating homonymous sense pairs than polysemous ones.
%Our work serves as evidence of the utility of vector space representations for capturing additional structure in the lexicon, and may offer improvements over discrete, symbolic ontologies that have until recently dominated research on the lexicon.
By demonstrating basic levels of consistency with human judgements, we hope to stimulate further research that combines contextualized word embedding techniques with discrete, symbolic ontologies to develop a more cognitively informed model of the lexicon.
\section*{Acknowledgements}
We would like to thank Steven Piantadosi, Ruthe Foushee, Sammy Floyd, and Jessica Mankewitz for thoughtful discussion of this work, and Jon Wehry for ensuring the experiment ran smoothly. We also thank the reviewers for helpful feedback, and the workshop organizers for giving us the opportunity to share this work.
%SN: this was the concluding sentence from the thesis, we should tie it to the lexicon at the end rather than language processing: Now that we demonstrate basic levels of consistency with human judgements, we can inspire further research to create formal, quantitative models \\
%for a critical problem in language acquisition and processing.\\

\newpage
\bibliographystyle{coling}
%\begin{footnotesize}
\bibliography{coling2020}
%\end{footnotesize}
\newpage
\appendix
\begin{appendices}
\section{Experiment Details}
\begin{figure}[h]
    \centering
    \includegraphics[scale = 0.23]{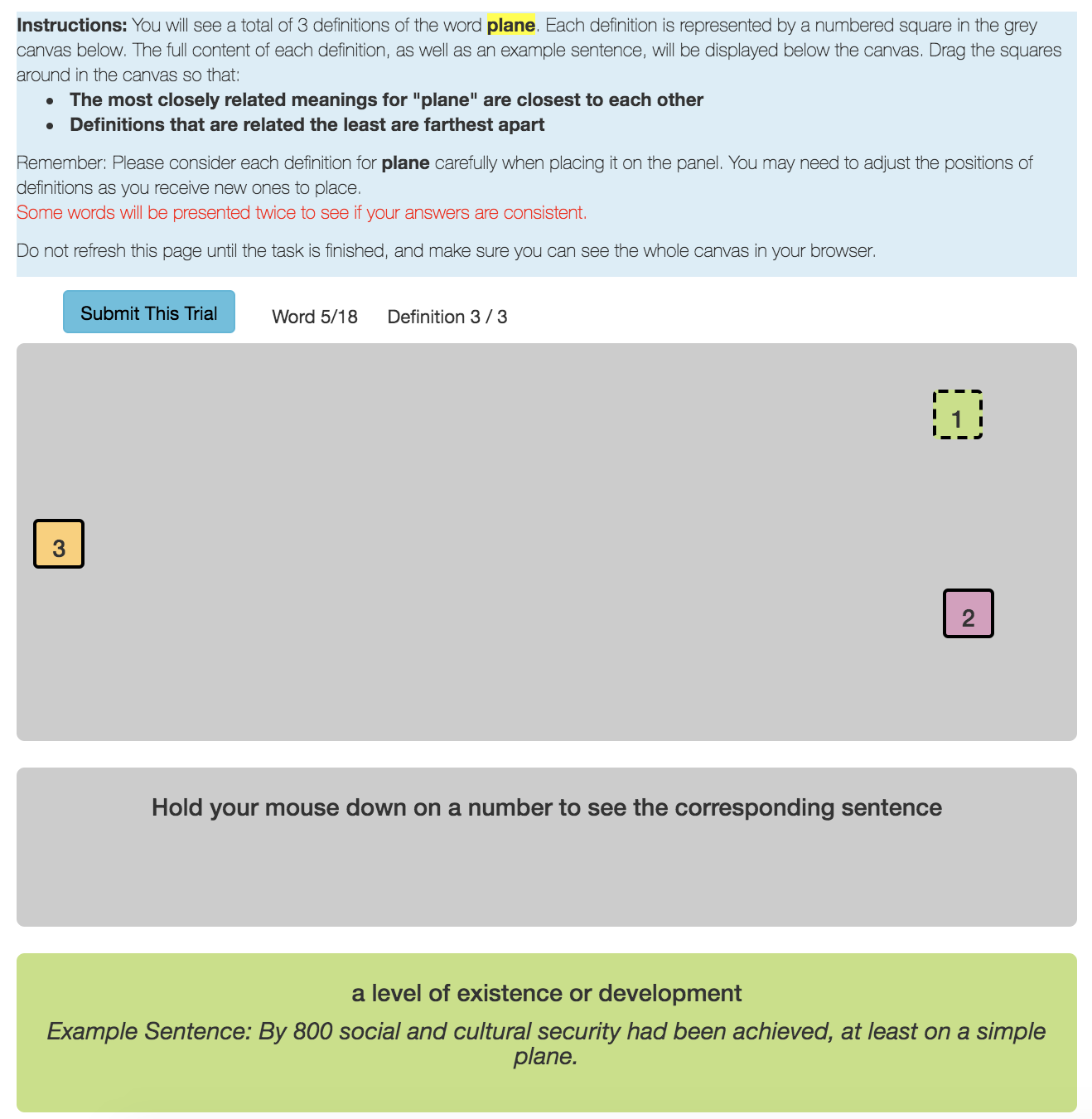}
    \caption{A sample trial in the experiment}
    \label{fig:expt_ui}
    %\vspace{-9mm}
\end{figure}
\section{Trial Types and Exclusion Criteria}\label{trialtypes}
The first two of the eighteen trials participants received were presented as \textit{training} trials to ensure that participants were familiar with the interface; data from these trials were discarded.
%The word types for these trials were \texttt{bank.n} and \texttt{bass.n}, and participants placed two polysemous senses and one homonymous sense for both trials on the canvas.
Participants received a mixture of \textit{shared} trials completed by all participants, and \textit{test} trials where only a subset of participants provided judgments.
The inclusion of both trial types gave us a consistent set of word types for which all participants contributed data (allowing us to evaluate which participants gave unreliable responses compared to all other participants), while at the same time characterizing a broad set of word types with more sparse responses.
Test trials were drawn from the set of 26 lemmas consisting of words from \newcite{SrinivasanRabagliati15}. 
To identify participants who provided low-quality data, we computed hold-one-out correlations for each participant using their relatedness matrices for the shared trials, and excluded participants whose data had a rank correlation with the hold-one-out averages that was lower than 0.4.
This threshold corresponds roughly to the 92nd percentile of scores if sense tokens are placed randomly in the interface.%delete poor quality
%Participants with rank correlations lower than 0.4 with the hold-one-out averages were excluded from analysis.

In addition to the shared and test trials, participants also saw two \textit{repeat} trials, drawn from the same set as the test trials, to evaluate the reliability of their responses within the same testing session.
Participants with rank correlations in reported distances between their original and repeat trials lower than 0.2 were excluded from analysis, corresponding to the 70th percentile of correlations if sense tokens are placed randomly (this threshold was lower because of variation in both the number of senses and the recency of the test trial). 
Ten participants who failed to meet both criteria were excluded from further analysis.
%Because self-consistency and group-consistency filters were applied concurrently, 10 participants were excluded from further analysis.

%\subsubsection{Aggregate Relatedness Matrices}
\end{appendices}

\end{document}